\documentclass[letterpaper]{article} 
\usepackage{aaai2027}  
\makeatletter
\usepackage[hyphens]{url}
\g@addto@macro\@maketitle{%
  \begin{center}
    \includegraphics[width=\textwidth]{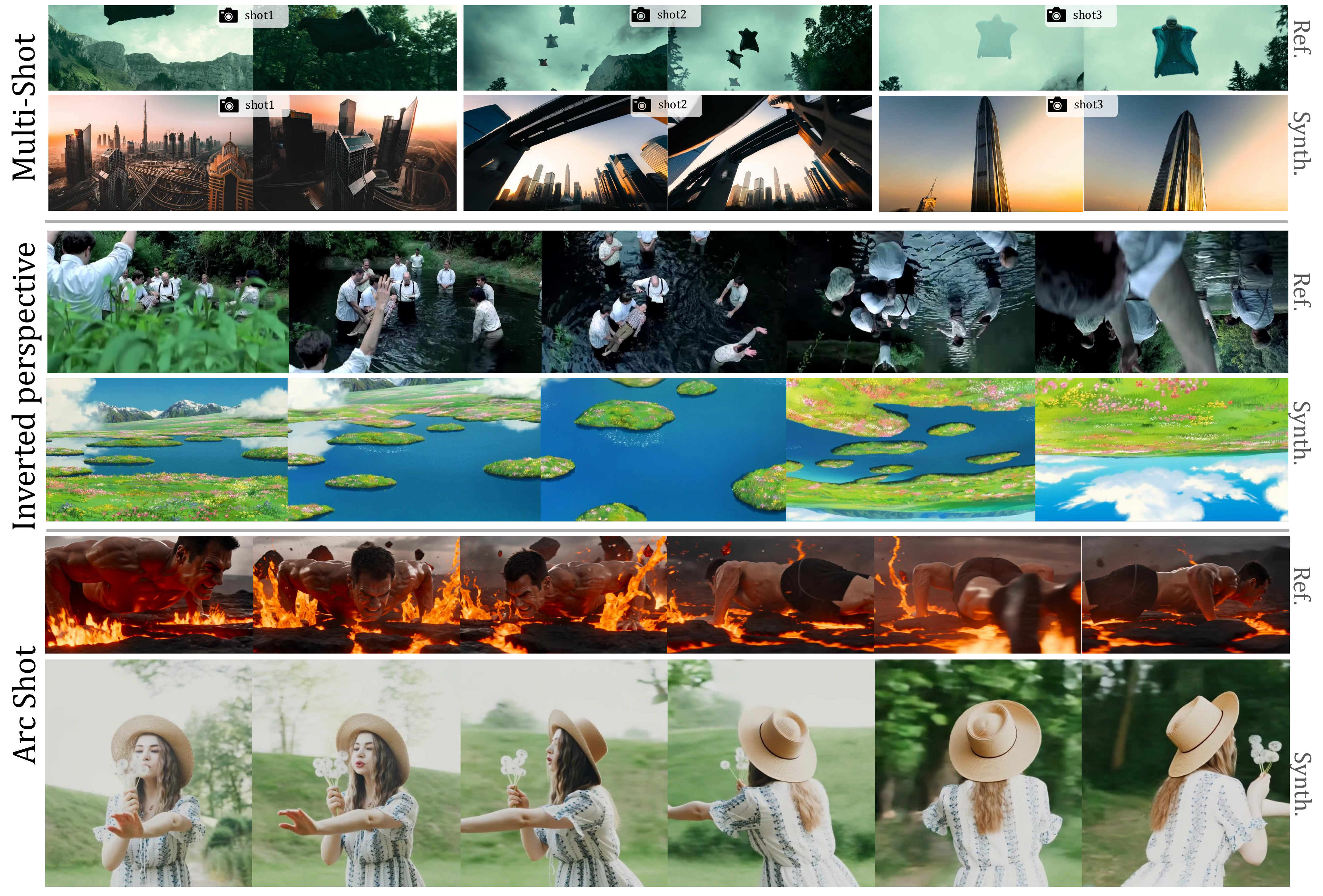}%
    \captionof{figure}{OmniDirector precisely clones diverse camera motions from reference videos to animate source images, remaining invariant to content discrepancies, aspect ratios, and spatial scales.}
    \label{topview} %
  \end{center} %
}

\makeatother

\usepackage{graphicx} 
\usepackage{natbib}  
\usepackage{caption} 
\usepackage{algorithm}
\usepackage{algorithmic}

\usepackage{amsmath}
\usepackage{amssymb}
\usepackage{multirow}

\usepackage{newfloat}
\usepackage{listings}
\DeclareCaptionStyle{ruled}{labelfont=normalfont,labelsep=colon,strut=off} 
\floatstyle{ruled}
\newfloat{listing}{tb}{lst}{}
\floatname{listing}{Listing}

\usepackage{booktabs}

\title{OmniDirector: General Multi-Shot Camera Cloning without Cross-Paired Data}

\author{
    Jiwen Liu\textsuperscript{\rm 1}\equalcontrib, Shujuan Li\textsuperscript{\rm 1,\rm 2}\equalcontrib, Zhixue Fang\textsuperscript{\rm 1}, Xiaohan Li\textsuperscript{\rm 1}, Yan Zhou\textsuperscript{\rm 1}\corresponding, Zijie Meng\textsuperscript{\rm 1,\rm 3}, \\ Zhimin Zhang\textsuperscript{\rm 1,\rm 3}, Yawen Luo\textsuperscript{\rm 1}, Guoxin Zhang\textsuperscript{\rm 1}, Yu-Shen Liu\textsuperscript{\rm 2}, Pengfei Wan\textsuperscript{\rm 1}
}
\affiliations{
    \textsuperscript{\rm 1}Kling Team, Kuaishou Technology, 
    \textsuperscript{\rm 2}Tsinghua University,
    \textsuperscript{\rm 3}Peking University\\
} 

\begin{document}
\maketitle
\begin{abstract}
\vspace{-0.3cm}
Cloning camera motion from reference videos is an important task in video generation, as videos provide intuitive and precise control. Existing methods either directly use parametric representations that fail to handle multi-shot generation or synthesize cross-paired data, which suffer from data scarcity, resulting in poor performance in complicated camera motion cloning.
To address these issues, we introduce a general camera motion representation that encodes cameras as grid motion videos. This camera grid represents the camera parameters visually and supports the integration of diverse trajectories for multi-shot video generation.
Building upon this, we propose OmniDirector, a unified framework trained on a million-scale camera grid-video pairs that coordinates characters, actions, and cameras to provide director-level control for multimodal diffusion transformers. 
Furthermore, we design a novel hierarchical prompt expansion agent that harmoniously integrates different control signals by systematically describing camera motion and visual content through understanding signal relationships. Extensive experiments demonstrate the superior performance and outstanding controllability of our framework. Project page:  \textcolor{blue}{{\url{https://ymlinfeng.github.io/OmniDirector.github.io/}}}%

\end{abstract}

\section{Introduction}
Camera motion is a vital component of video generation, which transcends spatial composition to shape the atmosphere, emotion, and narrative depth.
Conventional approaches to camera control in video generation typically employ either textual descriptions~\cite{wang2026multishotmaster,luo2026shotstream} or explicit camera parameters~\cite{he2024cameractrl, bai2025recammaster,wang2025cinemaster,meng2026make}. However, these modalities present an inherent trade-off: textual specifications fail to precisely define nuanced cinematographic attributes, while parametric representations impose significant user barriers to accessibility. Consequently, directly cloning camera motion from reference videos emerges as an optimal representation that reconciles both precision and accessibility requirements. Particularly, multi-shot camera cloning remains an under-explored problem.

Video reference-based camera control methods for video generation can be primarily categorized into explicit and implicit approaches. Explicit methods~\cite{bai2025recammaster, he2024cameractrl, wu2025cinetrans, meng2026argusstackedmultiviewidentity} typically inject camera motion into video generation models using matrices~\cite{bai2025recammaster} or Plücker coordinates~\cite{he2024cameractrl}. However, these frameworks are limited to representing basic camera motions and fail to handle shot transitions in multi-shot videos. Furthermore, the inherent semantic gap between these parameterized representations and standard visual signals (such as images and videos) complicates the model optimization. 
Implicit methods~\cite{luo2025camclonemaster} directly encode camera information from paired videos that differ solely in camera motion, yet such strictly controlled data is exceedingly scarce in real-world scenarios. Although recent methods~\cite{bai2025recammaster, luo2025camclonemaster} alleviate this issue with synthetic data, these methods fail to handle complicated cameras due to the limited diversity. Additionally, these videos inherently contain substantial extraneous information (such as appearance, character motion), frequently resulting in information leakage during practical deployment. 

To address these challenges, we introduce the camera grid, a general camera representation that handles complex camera motions for single or multi-shot videos in a unified way. Specifically, we first extract camera parameters from reference videos, then render them as a grid motion video which shows the movement within a 3D empty scene. This representation offers three key advantages: (1) \textbf{Generality}: The camera grid enables unified handling of diverse camera motion for multi-shot cloning. (2) \textbf{ Decoupling}: The empty scene decouples camera motion from other signals, preventing interference information from being injected into the model. (3) \textbf{Scalability and Compatibility}: The camera grid facilitates diffusion models' learning of camera motion from both data and optimization perspectives. For data, any video sample can generate its corresponding camera grid, facilitating construction from massive internet-scale datasets. For optimization, the camera grid is modality-compatible with other visual signals (e.g., images and videos), making it easier for video diffusion models to interpret.

Based on the camera grid, we propose a novel framework OmniDirector, a unified framework designed to provide director-level control for Multi-Modal Diffusion Transformers (MMDiTs)~\cite{esser2024scaling}. We train the model on a newly curated dataset of million-scale camera grid-video pairs. By utilizing the camera grid as an optimization-friendly visual-semantic representation, our approach effectively bridges the modality gap and facilitates camera motion learning from both data and optimization perspectives.

Furthermore, to harmoniously integrate camera control with other control signals into a unified whole, we design a novel hierarchical Prompt Expansion (PE) Agent mechanism during the inference stage. First, we design a camera prompt generator to describe the camera motion. We decompose the camera prompt into two hierarchies: inter-shot and intra-shot. The inter-level prompt handles the shot relationships across shot transitions to ensure semantic coherence in multi-shot scenarios. The intra-level focuses on the camera clone descriptions within individual shots. Finally, through semantic fusion, we integrate camera motion, subjects, and object motion into a unified representation.
Overall, our work makes the following key contributions: 
(1) We propose OmniDirector, a unified video generation framework that achieves general multi-shot camera cloning without cross-paired data, empowering MMDiTs with director-level control.
(2) We introduce the camera grid, a general representation of various camera movements for multi-shot camera cloning, and construct a million-scale camera grid–video dataset to scale up the training of our model.
(3) At inference, the hierarchical prompt expansion agent integrates camera motion with other control signals into a harmonious text, enabling collaborative creation with multimodal signals.
\vspace{-0.5cm}
\section{Related Work}
\subsubsection{Video Generation.} Video generation has advanced significantly, driven by the success of Diffusion Models~\cite{ho2020denoising, ho2022video, ho2022imagen}. Based on input conditions, the field is primarily categorized into Text-to-Video (T2V)~\cite{singer2022make, villegas2023phenaki, bar2024lumiere, polyak2024movie}, Image-to-Video (I2V)~\cite{zhang2023i2vgen, xing2024dynamicrafter, chen2023seine, renconsisti2v}, and Video-to-Video (V2V)~\cite{wang2023videocomposer} tasks. T2V focuses on complex semantic-to-motion translation; I2V emphasizes temporal extrapolation while preserving spatial identity; and V2V targets attribute manipulation while maintaining structural priors.
Architecturally, early diffusion-based video generation models predominantly relied on 3D U-Nets~\cite{ho2022video, ho2022imagen, singer2022make, blattmann2023stable} equipped with factorized spatial-temporal attention to model inter-frame dynamics. However, driven by scaling laws, the paradigm has rapidly shifted towards Diffusion Transformers (DiT)~\cite{peebles2023scalable, malatte}. To further enhance condition adherence, the Multi-Modal Diffusion Transformer (MMDiT)~\cite{esser2024scaling} has emerged as a leading framework. By decoupling text and visual tokens into separate streams and enabling interaction via joint attention blocks, MMDiT achieves superior cross-modal alignment~\cite{polyak2024movie}. This architectural evolution significantly improves scalability, enabling the generation of longer, high-fidelity, and physically plausible videos. 

\subsubsection{Camera Controllable Video Generation.} Despite the remarkable success of text-to-video generation models, relying solely on text prompts falls short in providing the precise spatial-temporal control required for real-world applications~\cite{zhang2023adding, mou2024t2i,chen2023control}. Consequently, introducing additional conditional signals for controllable generation has been widely studied~\cite{zhao2024motiondirector, guo2023animatediff, hu2024animate, yin2023dragnuwa, chen2024videocrafter2, girdhar2024factorizing}. Among various control modalities, camera movement stands out as a fundamental cinematic language. Existing camera-controllable methods~\cite{xu2024camco, zheng2024cami2v, bahmani2025ac3d} can be broadly categorized by their reliance on explicit camera parameters.
The first category requires explicit parameters to dictate camera dynamics. For instance, MotionCtrl \cite{wang2024motionctrl} encodes 6DoF camera extrinsics and injects them into temporal attention layers, while CameraCtrl \cite{he2024cameractrl} employs Plücker embeddings and a dedicated camera encoder to capture richer geometric information. To further enforce 3D geometric constraints, methods like CamCo \cite{xu2024camco} and CamI2V \cite{zheng2024cami2v} utilize epipolar attention. Recently, AC3D \cite{bahmani2025ac3d} conducted an in-depth investigation into camera motion within diffusion transformers, achieving enhanced visual quality. While these methods achieve precise viewpoint control, a key limitation is that obtaining and specifying explicit camera trajectories can be cumbersome for general users.
To alleviate this, the second category explores parameter-free or training-free approaches. AnimateDiff \cite{guo2023animatediff} introduces various motion LoRAs to learn specific patterns of camera movements without requiring frame-wise parameters. Furthermore, methods like MotionMaster \cite{hu2024motionmaster} and MotionClone \cite{ling2024motionclone} employ an inversion process to derive motion representations directly from temporal attention maps. Although these parameter-free methods offer greater user convenience, they often exhibit limited generalization and can struggle to maintain robust control in complex scenarios.
\vspace{-0.3cm}

\begin{figure}[t]
\begin{center}
\includegraphics[width=1.0\linewidth]{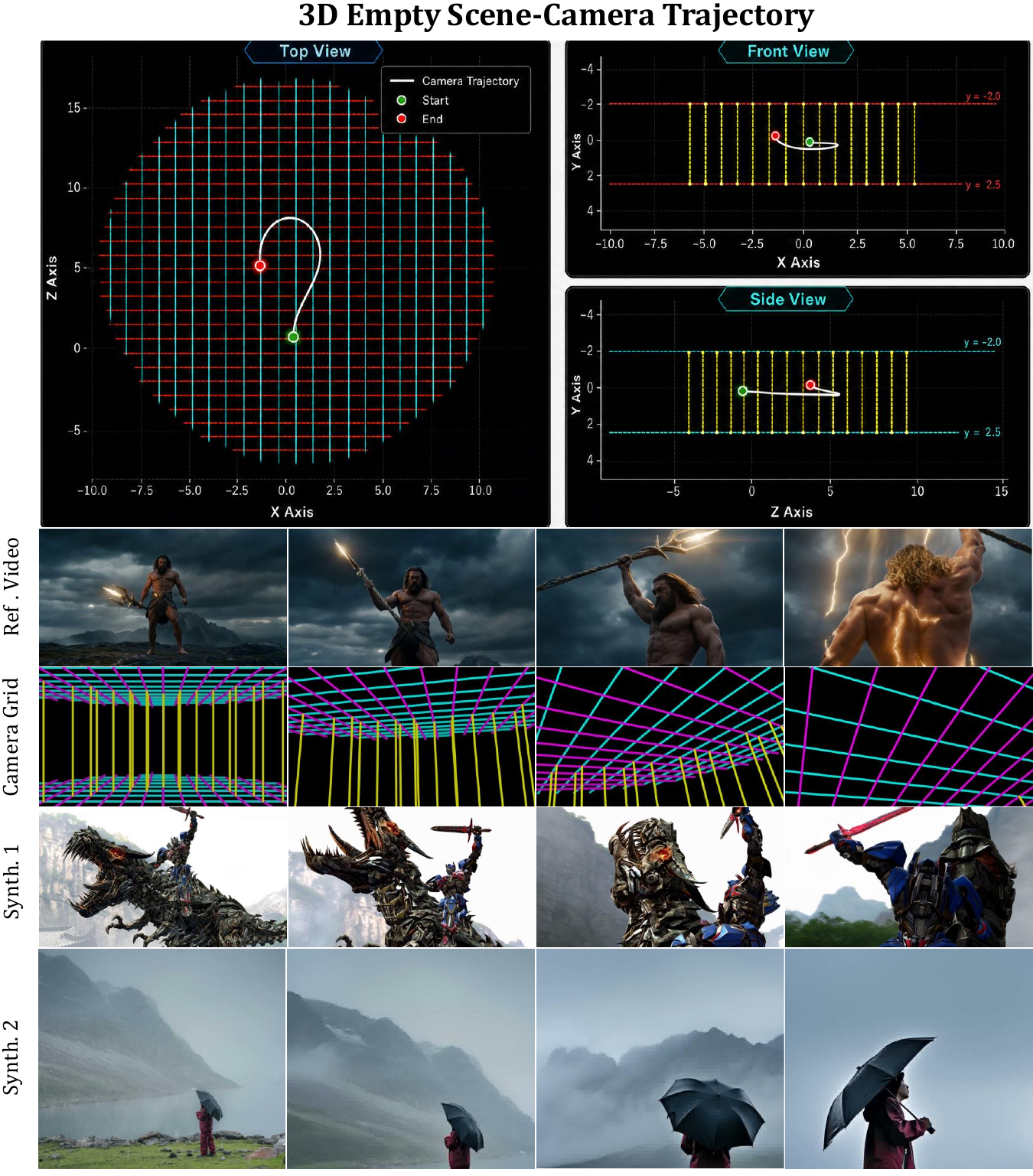}
\end{center}
\caption{\textbf{3D Scene Modeling in Camera Grid}. \textbf{Top}: Given reference camera poses, we simulate spatial motion within an empty 3D scene. Orthogonal lines represent the ceiling and floor (red and blue), while vertical lines (yellow) denote the walls. \textbf{Bottom}: Rendering this grid scene from varying viewpoints yields the camera grid, providing a visual representation of camera motion that conditions the model to generate videos with similar trajectories.}
\label{fig:cam_render}
\end{figure}

\subsubsection{Video-referenced Camera Controllable Video Generation.} To bridge the gap between the precise but cumbersome explicit parameter control and the user-friendly but coarse text-driven generation, video-referenced camera control has emerged as a highly practical paradigm. By leveraging a source video as a motion exemplar, this approach offers superior precision over pure text prompts while remaining significantly more intuitive for users than manually specifying complex trajectories. Existing methods in this domain can be broadly classified into two categories. The first category adopts a parameter-extraction pipeline~\cite{wang2024motionctrl, he2024cameractrl, xu2024camco, zheng2024cami2v}, where explicit camera parameters are estimated from the reference video and subsequently injected into the generation model as conditions. However, this paradigm is fundamentally bottlenecked by inherent scale ambiguities across different scenes. Applying camera translations extracted from a reference environment to a generated scene with a disparate spatial scale often leads to severe geometric distortions and unnatural dynamics.
The second category bypasses explicit parameter extraction by directly training models on cross-paired data—pairs of videos that share identical camera movements but depict different contents. While this end-to-end approach \cite{luo2025camclonemaster, ling2024motionclone, hu2024motionmaster} mitigates scale mismatch, it is severely constrained by the scarcity of real-world paired data, hindering the ability to scale up model training. To overcome this data scarcity, some recent methods~\cite{li2024director3d, he2025cameractrlii} resort to synthetic datasets rendered via game engines like Unreal Engine. Nevertheless, these synthetic videos typically lack the rich narrative contexts found in real-world footage, making it exceedingly difficult for models to comprehend and handle complex cinematic signals, such as abrupt shot cuts and intricate scene transitions.
\vspace{-0.5cm}

\section{Method}

In this section, we detail the design of our proposed OmniDirector in Figure \ref{fig:overview}. We first describe the core representation camera grid. Next, we introduce OmniDirector’s architecture and training strategy. Finally, we introduce the hierarchical prompt expansion agent and inference strategies.

\begin{figure}
\begin{center}
\includegraphics[width=1.0\linewidth]{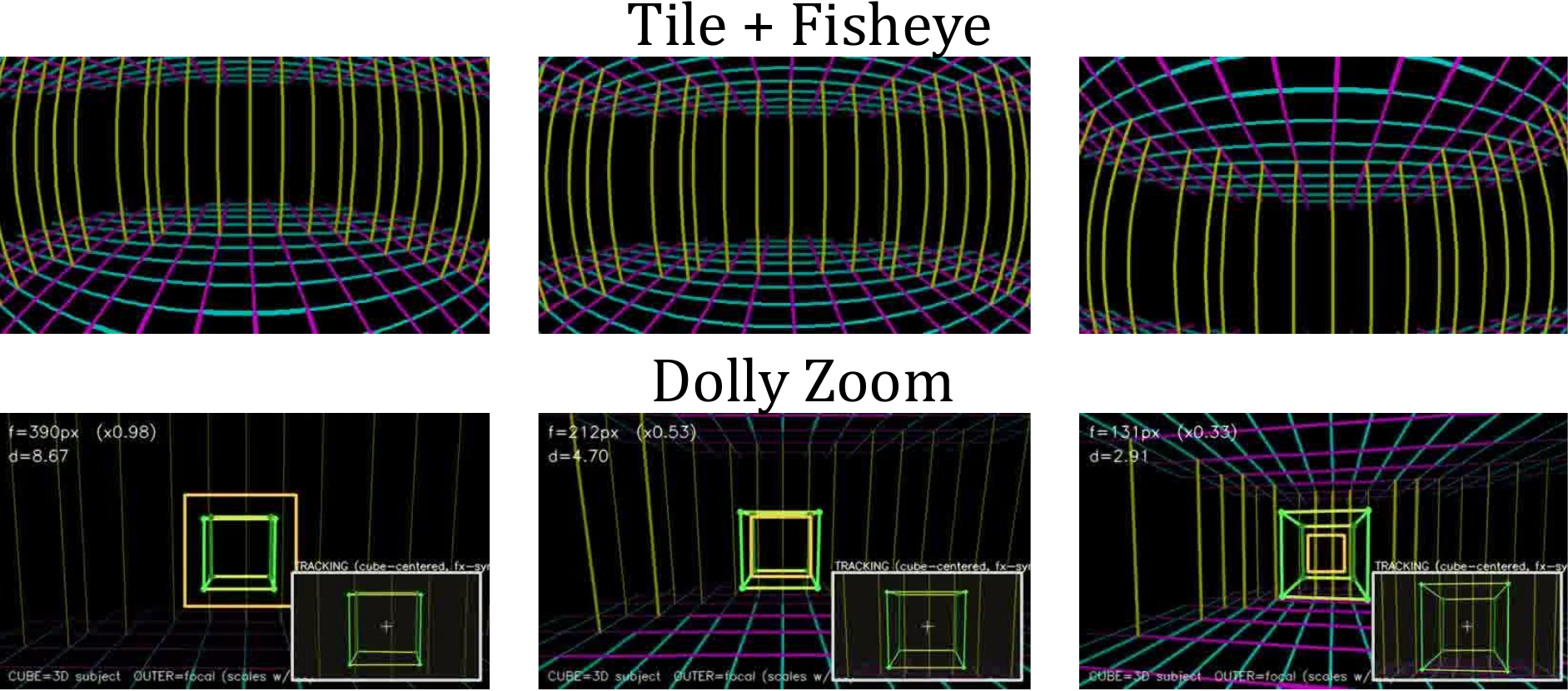}
\end{center}
\caption{Extension of the camera grid to special camera effects. \textbf{First row:} fisheye distortion, where straight lines are rendered as smooth curves via the Kannala--Brandt model. \textbf{Second row:} dolly zoom (Hitchcock zoom), where the subject remains fixed in size while the background undergoes pronounced perspective stretching.}
\label{fig:extension_of_camera_grid}
\vspace{-0.5cm}
\end{figure}

\subsection{Camera Grid}
Camera movement is the dynamic change of viewpoint over time, which reveals the spatial relationships among objects within a scene.
To accurately simulate the spatial transformation with camera movement, we adopt a simplified modeling approach: we abstract the complex real world into an empty room as shown in Figure~\ref{fig:cam_render}, where no objects or scene elements are placed. Only 3D grid lines are used to indicate the directions of spatial coordinate axes, thereby clearly presenting the geometric structure of the space and the camera motion trajectory.

\subsubsection{Visualizing the camera moving in an empty scene.}
Given a sequence of camera matrix parameters $P=\{R_i, t_i\}^T_{i=1}$ from the reference video, where $R_i \in SO(3)$ denotes the rotation matrix, $t_i \in \mathbb{R}^3$ represents the translation vector, and $T$ denotes the total number of video frames.
First, we compute the spatial bounding box of the scene by analyzing the camera translation trajectory $\{t_i\}$. In the scene box, we use two planes to represent the floor and ceiling. Specifically, uniformly sampled grid points $\{p\}$ are generated on the X-Z plane with two fixed heights. We define the heights as relative to the average scene height $\overline{y}$ along the Y-axis with an offset $\Delta h$ as follows:
\begin{equation}
 y_{floor} = \overline{y} - \Delta h
\end{equation}
\begin{equation}
 y_{ceiling} = \overline{y} + \Delta h
\end{equation}
In our experiments, $\Delta h $ is defined proportionally to the median distance between consecutive camera poses, with a lower bound to ensure robustness. Orthogonal grid lines are then rendered on these two planes to form a spatial reference framework, as shown in Figure \ref{fig:cam_render}.

\begin{figure*}[t]
\begin{center}
\includegraphics[width=1.0\linewidth]{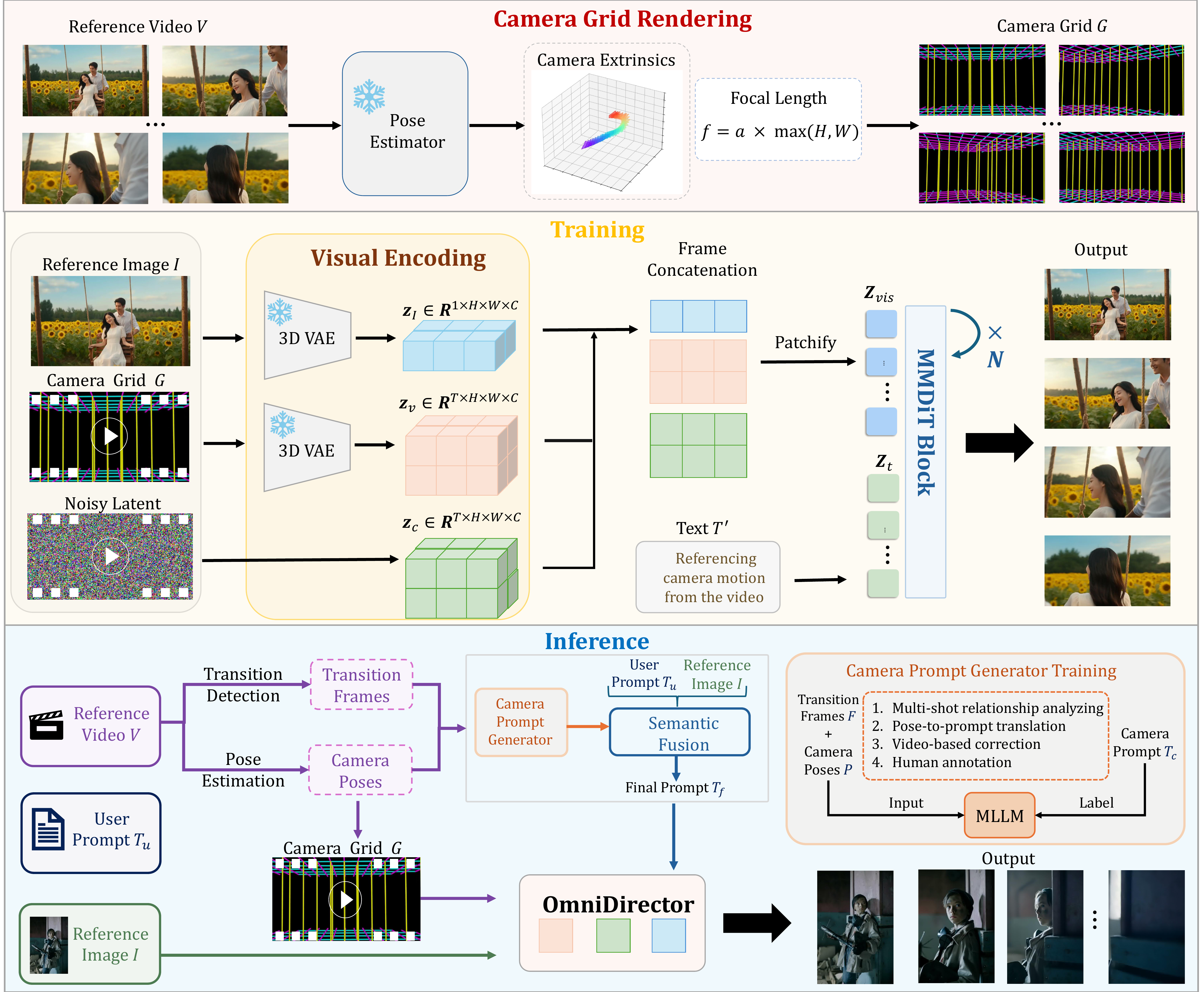}
\end{center}
\vspace{-2.0mm}
\caption{\textbf{Overview of OmniDirector}. 
\textit{\textbf{Top}}: OmniDirector represents camera motion via a camera grid $G$, which is obtained by rendering the camera poses of a reference video $V$ as movement within an empty 3D space. \textit{\textbf{Middle}}: During training, the camera grid is injected into the MMDiT alongside other control signals via token concatenation. \textit{\textbf{Bottom}}: At inference, a PE Agent harmoniously integrates various signals into the text prompt, achieving unified multi-signal control.}
\label{fig:overview}
\vspace{-6mm}
\end{figure*}

To enhance the spatial perception of camera motion, we use lines along the Y-axis to present the relative movement of objects. For the sake of computational efficiency and visual simplicity, rather than connecting all grid points between the two X-Z planes, we construct a tubular boundary structure surrounding the trajectory. Specifically, we project the camera trajectory onto the X-Z plane, yielding projected points $c_{proj}$. A KD-tree is employed to compute the distance $d_{traj}$ from each grid point $p$ on the X-Z plane to the projected points $c_{proj}$, and vertical line segments are generated within an annular region:
\begin{equation}
W = \{(x, z) \mid r < d_{traj}(x, z) < r + \delta \}
\end{equation}
where $r$ denotes the inner tunnel radius and $\delta$ represents the wall thickness. These vertical lines connect the floor and ceiling planes, creating a visual ``tunnel wall'' effect. 

After completing the spatial scene modeling, spatial variations are achieved by rendering views under different camera poses. For camera movements (such as dolly, zoom, pan, and tilt), during the rendering of each frame, the grid line segment endpoints $\{P_w\}$ in the world coordinate system are first transformed to the camera coordinate system via the camera extrinsic matrix $[R_i \mid t_i]$ as $P_c = R_iP_w + t_i$, and then mapped to the image plane through projection transformation using the camera intrinsic parameters. Finally, we synthesize temporal camera motion by rendering the per-frame poses.

\subsubsection{Extending the camera grid to represent special camera effects.}
The camera grid can be extended to represent certain special camera effects by modifying its rendering scheme. As illustrated in the Figure~\ref{fig:extension_of_camera_grid}, we take fisheye distortion and dolly zoom as two representative examples.

\textbf{Fisheye Distortion.}
For fisheye distortion, we adopt a distortion formulation based on the Kannala--Brandt model~\cite{kannala2006generic}. The core procedure first computes the incident angle of a spatial point with respect to the camera's optical axis,
\begin{equation}
    \theta = \arctan(r' / \zeta),
\end{equation}
where $\theta$ denotes the incident angle between the line of sight to the spatial point and the optical axis, $\zeta$ is the depth of the point along the optical axis (i.e., its $z$-coordinate in the camera coordinate system), and $r'$ is the radial distance of the point from the optical axis on the plane perpendicular to it. A fourth-order polynomial is then applied to perform radial nonlinear scaling of this angle, yielding the distorted angle
\begin{equation}
    \theta_d = \theta\left(1 + k_1\theta^2 + k_2\theta^4 
    + k_3\theta^6 + k_4\theta^8\right),
\end{equation}
where $\theta_d$ is the distorted (mapped) angle, and 
$k_j, j\in\{1,2,3,4\}$ are the radial distortion coefficients of the 
Kannala--Brandt model that governs the nonlinear projection behavior of the fisheye lens. In the rendering implementation, each 3D spatial line is densely subdivided into a set of discrete points, and the aforementioned nonlinear projection mapping is applied to each point individually. In this way, the originally straight physical lines are fitted into smooth curves on the pixel plane, thereby faithfully reproducing the characteristic ``straight-line-to-curve'' visual distortion inherent to fisheye lenses.

\textbf{Dolly Zoom.}
The dolly zoom effect is represented by constructing a 3D cube at the subject's location in conjunction with a picture-in-picture (PIP) tracking view. Its core mechanism exploits the principle that the focal length is proportional to the distance,
\begin{equation}
    \varphi \propto \rho,
\end{equation}
where $\varphi$ denotes the focal length of the camera and $\rho$ is the distance from the camera to the subject. This proportionality ensures that the projected visual size of the 3D cube remains constant. Simultaneously, a 2D bounding frame scaled by the focal-length ratio $\varphi / \varphi_{\mathrm{ref}}$ is rendered to quantify the change in field of view, where $\varphi_{\mathrm{ref}}$ denotes the reference focal length serving as the baseline for the scaling. In parallel, a virtual 
tracking camera generates the picture-in-picture view in which the 
subject is always kept centered; by leveraging the drastic perspective deformation of the background tunnel grid during zooming, this design precisely reproduces the signature ``static subject, stretching background'' visual characteristic of the dolly zoom effect.

\subsubsection{General camera representation for multi-shot camera cloning.}
As a purely visual representation, the camera grid readily generalizes to encode camera motion across multi-shot video sequences.

Given a reference video $V$, we first adopt the transition detection model TransNet-V2~\cite{soucek2024transnetv2} to detect the shot transition frames $F=\{f_i\}_{i=1}^{K}$, where $K$ is the total number of transition frames. We segment the video into distinct sub-clips based on $F$ to ensure scene consistency within each segment. We render the transition frames as special pure white frames to construct a transition signal. 

Then we treat each sub-clip as a single shot video. We leverage DPA-V3 \cite{lin2025dpa3} to estimate camera extrinsic parameters, and approximate the camera focal length with $f=a\times \text{max}(H, W)$ to enhance the stability of camera control and ensure resolution-agnostic generalization, where $a=0.8$ is an empirical scale factor. Then we render the camera grid $G$ with the process in the previous section.

By naturally extracting a corresponding camera grid from arbitrary video samples, we can automatically construct paired training data consisting of reference videos and their camera grids. This inherent capability means that the generation of these data pairs can be seamlessly scaled up, facilitating the efficient curation of massive, internet-scale datasets for robust model training.

\subsection{OmniDirector}

\subsubsection{Injecting camera grid via token concatenation.}
The proposed camera grid representation achieves complete disentanglement of camera control from other visual attributes, enabling interference-free injection into the transformer architecture. Benefiting from the fact that the camera grid is essentially a spatiotemporal signal similar to video, it can be encoded following the standard video processing pipeline. 

Specifically, we first encode the camera grid $G$ and reference image $I$ into latents using a pre-trained 3D-VAE $z_c = \varepsilon(G) \in \mathbb{R}^{T \times H \times W \times C}$ and $z_I = \varepsilon(I) \in \mathbb{R}^{1 \times H \times W \times C} $, where $H \times W$ denotes the frame resolution, and $C$ denotes the latent channel. We concatenate these visual modalities with the video noisy latent $Z_v$ along the frame dimension to form a unified spatiotemporal representation $z_{vis}=\text{Concat}(z_I, z_v, z_c) \in \mathbb{R}^{(2T+1) \times H \times W \times C}$, which is then mapped into token sequences through 3D convolutional layers:
\begin{equation}
\mathbf{Z}_{vis} = \text{Patchify}(\mathbf{z}_{vis}) \in \mathbb{R}^{N_{vis} \times D}
\end{equation}
where $N_{vis}$ 
denotes the number of video tokens and $D$ is the hidden dimension. 
For the text condition $T'$, we process it through a pre-trained text encoder (e.g., T5): $\mathbf{Z}_t = \text{TextEncoder}(T') \in \mathbb{R}^{N_t \times D}$. 

Subsequently, we perform joint attention between the visual tokens and text tokens. The visual and text modalities are processed through separate attention pathways within the MMDiT architecture, where they interact through attention. Within each Transformer block, tokens from both modalities exchange information:
\begin{equation}
\mathbf{Z}_{vis}^{(l+1)} = \text{FFN}(\text{Attention}(\text{LN}(\mathbf{Z}_{vis}^{(l)}), \mathbf{Z}_t^{(l)})) + \mathbf{Z}_{vis}^{(l)}
\end{equation}
where $l$ denotes the layer and FFN denotes Feed-Forward Network. 

This hierarchical fusion design first unifies all visual modalities (reference image, video, and camera grid) into a coherent spatiotemporal representation, then enables text semantics to modulate the visual generation process through joint attention. This architecture maintains the natural alignment between visual signals while allowing flexible text-based control, supporting seamless injection of multimodal conditions in an end-to-end manner.

\subsubsection{Training strategy.}
To balance controllability and generation capability, we incorporate a
self-reconstruction objective into $30\%$ of the training samples.
For these samples, instead of pairing the camera grid $G$ with a natural
video, we replace the target video with the camera grid itself, i.e., the
model is conditioned on $\mathbf{z}_c = \varepsilon(G)$ and trained to
reconstruct $\mathbf{z}_v = \varepsilon(G)$.
Since the camera grid contains no appearance or content cues, faithfully 
reconstructing it forces the model to parse the geometric structure and
temporal dynamics encoded in the grid, rather than treating it as a weak
auxiliary hint that can be partially ignored.
This auxiliary objective thus strengthens the model's understanding of the
camera trajectory representation and prevents it from overfitting to
spurious correlations in the camera-to-video mapping.
The remaining follows the standard camera-conditioned
video generation task, where the target video depicts natural content
consistent with the conditioning camera grid.

\subsection{Unified Multimodal Control Signals for Inference}
\subsubsection{Hierarchical prompt expansion agent.}

Video generation models are commonly pre-trained on text-to-video (T2V) objectives, making textual conditioning a primary factor that governs the generated content. Consequently, consolidating heterogeneous control signals (e.g., user prompts, reference-image cues, and camera grids) into a unified textual condition is a crucial step for eliciting reliable camera controllability.

A straightforward approach to obtain camera-related text from a reference video is to directly ask a multimodal large language model  (MLLM), to describe the camera motion. However, camera motion in real videos is tightly entangled with other factors such as subjects, actions, and backgrounds. The MLLM tends to incorporate these non-camera semantics into the description. When used at inference time, such entangled signals can conflict with the intended conditioning, thereby weakening the effective camera control. To address this issue, we propose a hierarchical prompt expansion agent capable of generating semantically coherent prompts free from background leakage, conditioned on the given inputs as shown at the bottom of Figure \ref{fig:overview}. We divide the text expansion process into two parts: camera prompt generation and multimodal signal fusion.

\textbf{Camera prompt generation.} We first train a generator to describe the camera motions with prompt $T_c$ directly from camera parameters, including the transition frames $F=\{f_i\}_{i=1}^{K}$ and camera poses $P$. To achieve this, we decompose the camera prompt into inter-shot and intra-shot to ensure consistency across the whole video.  Specifically, we prepare the training data with four fundamental stages: (1) analyzing the relationship between adjacent shots and generating an inter-shot prompt $T_s$; (2) deriving the pose prompt descriptions $T_p$ through segment-wise textual matching based on the camera poses; (3) correcting $T_p$ by leveraging visual cues from reference videos; and (4) performing rigorous manual refinement to ensure high data quality.

Given the transition frames $F=\{f_i\}_{i=1}^{K}$, we first extract several frames immediately before and after the transition frame $f_i$, and prompt an MLLM to analyze the relative relational changes. Finally, we aggregate all the shot transition relationships into a comprehensive description $T_s$.

Then we segment the video into distinct sub-clips based on transition frames. For each clip, we then perform camera pose analysis to derive camera motion descriptions via a text-matching approach. Specifically, we segment the full camera pose trajectory into multiple sub-clips. For each sub-clip, we compute the relative pose between the first and last frames: $\Delta P = [\Delta R \mid \Delta t] = P_0^{-1}P_t$, where $\Delta t$ and $\Delta R$ denote the translational and rotational increments, respectively. For translation, we identify the dominant motion axis with the largest absolute displacement in $\Delta t$, which determines the motion direction; we further discretize the motion speed into fast, normal, or slow based on the translation distance. For rotation, we convert $\Delta R$ into Euler angles and determine the rotation type by the dominant angular component.
Additionally, we define arc shot with an explicit rule: a sub-clip is classified as arc shot if the dominant translation axis is $x$ and the dominant rotation component is yaw:
\begin{equation}
    \text{Left Arc Shot}: \Delta \theta_{yaw} > 0 \text{ and } \Delta x < 0
\end{equation}
\begin{equation}
    \text{Right Arc Shot}: \Delta \theta_{yaw} < 0 \text{ and }  \Delta x > 0 
\end{equation}
Finally, we merge consecutive sub-clips with redundant or equivalent motion labels to form a compact, complete camera-motion description $T_p$. 

Recognizing that pose estimation often yields inaccurate results in complex videos, we leverage reference videos for further rectification. Specifically, we extract keyframes from the reference video $V$ and feed them alongside the pose prompt $T_p$ into Qwen3-VL~\cite{bai2025qwen3}. The model is then instructed to synthesize both modalities to generate an accurate and refined camera motion description. After combining the inter-shot prompt $T_s$ and refined intra-shot prompt with Qwen3-VL~\cite{bai2025qwen3}, we incorporate manual annotation to further enhance the accuracy and reliability of the generated camera motion descriptions $T_c$.

Based on the aforementioned data construction pipeline, we fine-tune the Qwen3-VL model using camera parameters as inputs and the final camera motion descriptions as labels.

\textbf{Semantic Fusion.} To effectively transfer the camera motions acquired in previous stages to novel scenes, we must consolidate multiple control signals into a unified entity. Specifically, we leverage Qwen3-VL as our multimodal engine to seamlessly integrate the existing pose descriptions $T_c$, the reference image $I$, and the user prompt $T_u$ into a comprehensive and cohesive representation.


\subsubsection{Adaptive Classifier-Free Guidance Strategy.}
In video generation with multi-modal control signals, providing informative conditional guidance is crucial for fully unleashing the model's generative capabilities. Standard Classifier-Free Guidance (CFG) typically interpolates between conditional and unconditional predictions. Considering the unique visual characteristics of the camera grid representation, we specifically tailor its unconditional branch: we set the visual unconditional input to a completely black background and explicitly incorporate the description of a ``completely static camera'' into the negative text prompt. 

Furthermore, camera motion is fundamentally a global spatial transformation that dictates the macroscopic geometric outlines and scene layout of each frame. Consequently, we introduce a coarse-to-fine denoising scheduling strategy. Specifically, we inject the camera grid features during the high-noise regime of the diffusion process to establish the global spatial structure, while introducing other control signals during the low-noise regime to refine local contents and semantic details.

\section{Experiments}
\subsection{Experimental Setups}
OmniDirector is trained on top of an in-house image-to-video diffusion backbone. For training, we curate a large-scale dataset of 1.8M internet videos spanning diverse domains (e.g., movies and advertisements). Each video is preprocessed by resizing to 480p. During training, the model is conditioned on the triplet $\{G, I, T'\}$. We train the model for 10k optimization steps using a learning rate of $5\times 10^{-5}$ and a batch size of 64. During the training phase, data augmentation is performed by applying random colors to the camera grid, and pose jitter is introduced to ensure robust performance.

\subsubsection{Evaluation Metrics.} 
To comprehensively evaluate the performance of OmniDirector, we conduct extensive experiments focusing on its camera control capabilities and overall visual quality. 
Regarding camera control, we employ DPA-V3 \cite{lin2025dpa3} to extract the camera pose trajectories from both the reference and generated videos. We then compute the scale-invariant relative pose errors, namely Relative Rotation Error (RRE) and Relative Translation Error (RTE). RRE is defined as the angular difference between the poses of reference videos and generated ones. RTE measures the directional error of the translation. To further evaluate the robustness of the predictions, we report R-Pre and T-Pre. R-Pre is defined as the percentage of predictions with a relative rotation error below $4^{\circ}$. Similarly, T-Pre denotes the proportion of predictions with a relative translation error below $20^{\circ}$. 

Since our data contains a large number of complex real-world scenes, pose estimation methods inherently suffer from errors. To evaluate the performance of OmniDirector more comprehensively and accurately, we utilize Gemini 3.1 Pro and GSB (Good / Same / Bad) to quantify the results.

To evaluate the accuracy of shot transitions, we define the success rate from two dimensions: (1) Tem-Pre (Temporal): Measures the temporal alignment of the transitions. A predicted transition is considered successful if its temporal error relative to the reference transition is less than 3 frames. This metric is evaluated using the TransNet-V2 \cite{soucek2024transnetv2} shot boundary detection model. (2) Sem-Pre (Semantic): Evaluates the semantic consistency of the transitions. For temporally aligned samples, a transition is deemed successful if the Gemini 3.1 Pro model verifies that its transition type matches the reference.

Furthermore, we utilize the Gemini 3.1 Pro model to quantify the extent of reference video leakage at both the frame and shot levels. And we conducted a GSB pairwise comparison with CamCloneMaster \cite{luo2025camclonemaster} to analyze performance across three key areas: camera, quality, and narrative to conduct a more comprehensive assessment. 

\begin{figure}
\begin{center}
\includegraphics[width=1.0\linewidth]{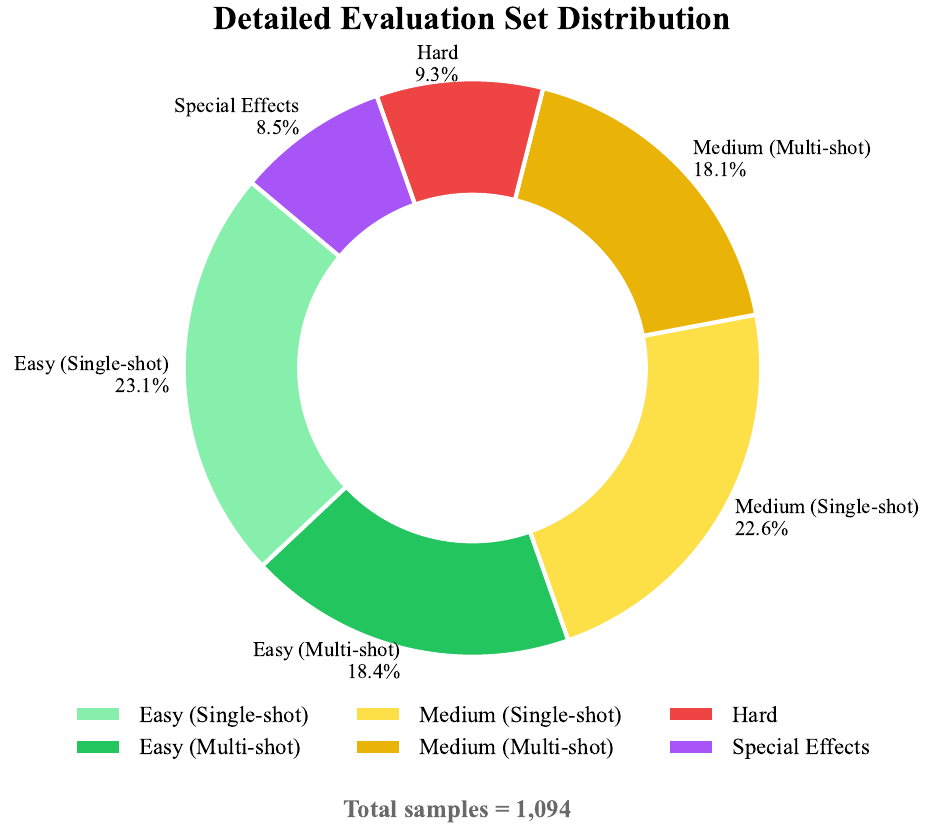}
\end{center}
\caption{The evaluation set distribution.}
\label{fig:data_distribution}
\end{figure}

\subsubsection{Evaluation Set.} We construct a validation set comprising 1,094 carefully curated samples, which are collected from the web, covering a wide range of domains including advertising, cinematic content, and several complex visual effects. This evaluation set encompasses a diverse range of scenarios, including in-domain and cross-domain reference video-image pairs, varying resolutions, single-shot and multi-shot sequences, as well as simple and complex camera trajectories. The detailed data distribution is shown in Figure \ref{fig:data_distribution}.

\subsection{Comparisons with State-of-the-Art Methods}
\subsubsection{Baselines.} We compare the proposed OmniDirector with state-of-the-art camera cloning method CamCloneMaster \cite{luo2025camclonemaster}, commercial models Seedance2.0 \cite{seedance2026seedance} and LTX-LoRA~\cite{cseti2024ltx23}. CamCloneMaster trains a Diffusion Transformer (DiT) architecture utilizing cross-paired data. Seedance2.0, representing the latest state-of-the-art commercial model, is evaluated under its omni-reference mode. Additionally, LTX-LoRA builds upon LTX-Video 2.3~\cite{hacohen2025ltx2}, employing Low-Rank Adaptation (LoRA) fine-tuning specifically for camera motion control. 

\begin{table*}[t]
\centering
\begin{tabular}{l|c|c|c|c|c|c|c|c}
    \toprule
    \multirow{2}{*}{Method} & \multicolumn{4}{c|}{\textbf{Camera Accuracy}} & \multicolumn{2}{c|}{\textbf{Transition Accuracy}} & \multicolumn{2}{c}{\textbf{Leakage Rate}} \\
    \cline{2-9}
    & RRE $^{\circ}$ $\downarrow$ & R-Pre $\%$ $\uparrow$ & RTE $^{\circ}$ $\downarrow$ & T-Pre $\%$ $\uparrow$ & Tem-Pre $\%$ $\uparrow$ & Sem-Pre $\%$ $\uparrow$ & Frame $\%$ $\downarrow$ & Shot $\%$ $\downarrow$ \\
    \midrule
    Seedance2.0 & 8.33 & 56.49 &49.98 & 29.07 & 4.17 & $-$ & 4.43&20.90 \\
    CamCloneMaster & 4.11 & 74.14 &27.45 & 52.21 & 2.20 & $-$ &1.60&11.59\\
    LTX-LoRA &5.67 & 66.34 & 26.96 & 52.07 & 38.94 & 29.55&15.04&56.52\\
    Ours &\textbf{2.64} & \textbf{83.18} & \textbf{16.84} & \textbf{72.74} & \textbf{96.52} & \textbf{83.79}&\textbf{0.51}&\textbf{3.38}\\
    \bottomrule
\end{tabular}
\caption{\textbf{Quantitative Evaluations.} $-$ denotes that this feature is not supported. In comparison, we achieve superior performance across all evaluation metrics, including camera accuracy, transition accuracy, and leakage rate, which demonstrates the effectiveness of OmniDirector. }
\label{pose_comparison}
\end{table*}

\begin{table*}
\centering
\begin{tabular}{l|c|c|c|c|c|c|c}
    \toprule
    \multirow{2}{*}{Settings} & \multicolumn{4}{c|}{\textbf{Camera Accuracy}} & \multicolumn{2}{c|}{\textbf{Transition Accuracy}} & \textbf{Leakage Rate} \\
    \cline{2-8}
    & RRE $^{\circ}$ $\downarrow$ & R-Pre $\%$ $\uparrow$ & RTE $^{\circ}$ $\downarrow$ & T-Pre $\%$ $\uparrow$ & Tem-Pre $\%$ $\uparrow$ & Sem-Pre $\%$ $\uparrow$ & Shot $\downarrow$\\
    \midrule
    w/o Semantic Fusion& 3.85&78.20&19.90&67.45&94.40&78.30&4.10\\
    w/o Trans PE& 2.71&81.50&17.10&71.25&93.35&38.45&3.45\\
    w/o AdaCFG & 4.15& 74.55&21.41&62.30&94.10&80.20&3.83\\
    Full &\textbf{2.64} & \textbf{83.18} & \textbf{16.84} & \textbf{72.74} & \textbf{96.52} & \textbf{83.79}&\textbf{3.38}\\
    \bottomrule
\end{tabular}
\caption{\textbf{Ablation studies on our strategies} }
\vspace{-0.5cm}
\label{tab:ablation}
\end{table*}

\begin{figure*}
\begin{center}
\includegraphics[width=1.0\linewidth]{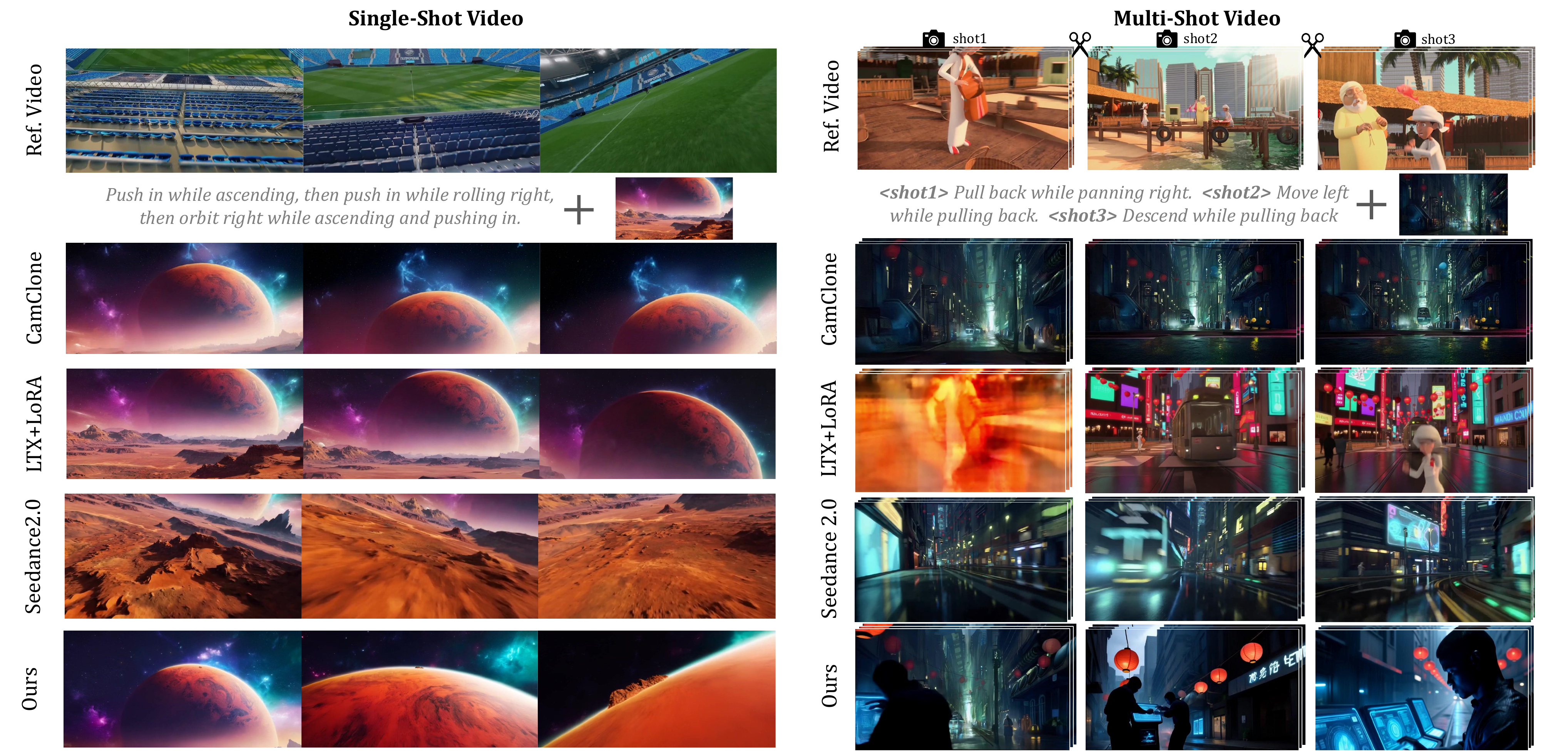}
\end{center}
\caption{\textbf{Qualitative Evaluations.} The results demonstrate that OmniDirector accurately clones the camera motion and shot transition semantics of the reference video.}
\label{fig:qualitative_comparison}
\end{figure*}

\subsubsection{Qualitative comparison.}
We illustrate the camera motion cloning capabilities of OmniDirector in Figures \ref{topview} and \ref{fig:cam_render}. As demonstrated, our model faithfully clones camera trajectories while exhibiting strong robustness against variations in scene scale, image resolution, and content discrepancy. Notably, in multi-shot scenarios, OmniDirector not only achieves accurate shot transitions but also strictly preserves the semantic coherence of the transitions.

We provide a qualitative comparison between OmniDirector and baseline approaches. As shown in Figure ~\ref{fig:qualitative_comparison}, Seedance2.0~\cite{seedance2026seedance} and CamCloneMaster~\cite{luo2025camclonemaster} struggle with multi-shot videos, yielding motion amplitudes that contradict human visual perception. While LTX-LoRA~\cite{cseti2024ltx23} manages to execute shot transitions, it suffers from substantial content leakage. Conversely, our proposed method seamlessly adapts to various complex scenarios, synthesizing high-quality videos that are highly consistent with the reference camera dynamics.

\subsubsection{Quantitative comparison.} We report the quantitative comparison results of camera control accuracy in Table~\ref{pose_comparison}. Our method outperforms all baselines across all evaluation metrics, demonstrating its effectiveness and its ability to clone cameras from reference videos accurately. Notably, we achieve a relative improvement of $39.3\%$ in translation precision (T-Pre) over the second-best method CamCloneMaster~\cite{luo2025camclonemaster}. This significant gain is because previous methods struggle with scale inconsistencies between the reference video and the source view, whereas our visual representation provides superior relative scale generalization. Furthermore, most existing methods fail to respond to reference videos with multiple shots, primarily due to the absence of such data in their training sets. While LTX-LoRA~\cite{cseti2024ltx23} can generate occasional shot transitions, the results in Table~\ref{pose_comparison} indicate that this capability is actually an artifact of information leakage rather than genuine camera control.

We quantify the leakage rates of different methods and present the results in Table~\ref{pose_comparison}. The results demonstrate that our approach achieves the lowest leakage rate compared to the baselines. This is because baseline methods rely heavily on the visual content of the reference video, whereas our camera grid representation and Prompt Expansion agent completely decouple the camera signal from the reference video. Notably, LTX-LoRA~\cite{cseti2024ltx23} exhibits the most severe leakage, which can be attributed to its insufficient number of fine-tuning parameters.

We report the GSB (Good/Same/Bad) comparison against CamCloneMaster~\cite{luo2025camclonemaster} in Table \ref{tab:gsb_comparison}. As illustrated, our approach demonstrates clear advantages across all three evaluated dimensions. These results strongly indicate that our method achieves superior performance in terms of both overall effectiveness and generation stability.

\begin{table}
    \centering
    \begin{tabular}{l|c|c|c}
    \hline
        &(G+S)/T $\%$& G/(G+B) $\%$ & (G+S)/(B+S) \\
        \hline
         Camera & 88.52 & 86.29 & 3.19\\
         Quality & 95.69 & 90.82 & 1.67 \\
         Narrative & 94.26 & 85.71 & 1.44\\
         \hline
         Average & 91.10 & 87.08 & 2.54 \\
    \hline
    \end{tabular}
    \caption{GSB comparisons of ours vs. CamCloneMaster. Notice that the average is weighted with $3:1:1$.}
    \label{tab:gsb_comparison}
    \vspace{-0.5cm}
\end{table}

\begin{figure*}
\begin{center}
\includegraphics[width=1.0\linewidth]{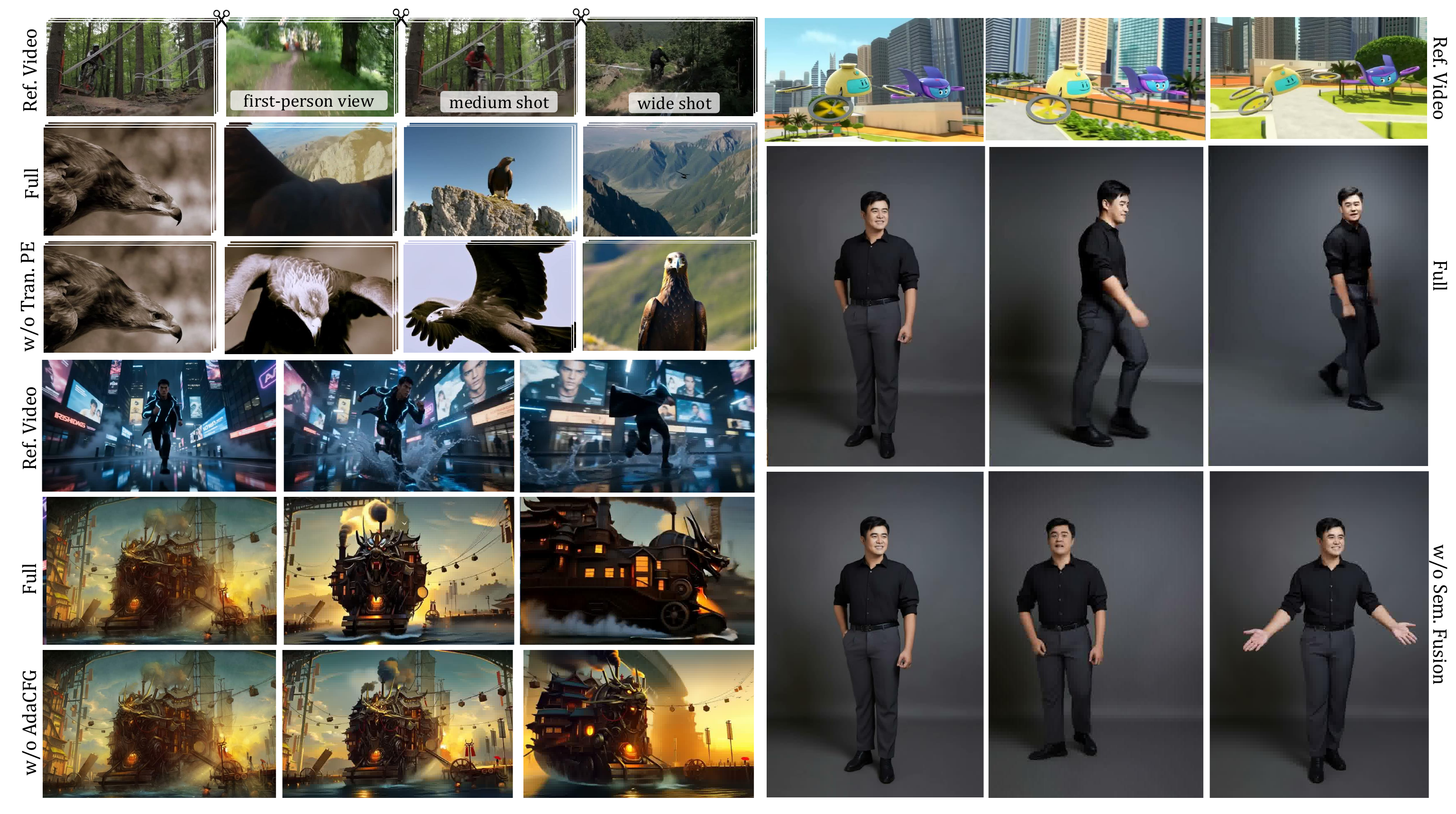}
\end{center}
\caption{Visualization of ablation studies.}
\label{fig:ablation_vis}
\end{figure*}

\subsection{Ablation study}

\subsubsection{The effect of PE agent design.} 
To validate the effectiveness of our design, we evaluate the prompts generated at different stages of the PE agent module during inference. As shown in Table \ref{tab:ablation}, comparing the ``w/o PE Semantic Fusion'' and ``Full'' configurations reveals that fusing multiple signals via the MLLM leads to comprehensive improvements across all metrics. Supported by the visual results in Figure \ref{fig:ablation_vis}, this multimodal signal fusion seamlessly integrates the camera motion with the reference image, yielding highly plausible visual content and camera trajectories. Furthermore, removing the fusion significantly increases the leakage rate. This is because the fusion effectively reduces conflicts by integrating multiple control signals.

The role of the inter-shot prompt is demonstrated in the ``w/o Trans PE'' row. Compared to the ``Full'' model, the absence of inter-shot relationship modeling degrades all metrics, with a particularly severe drop in the semantic precision of shot transitions (Sem-Pre). This occurs because, without guidance after a transition, the generated video shifts to a random scene that fails to correlate with the reference camera motion. Ultimately, our multi-stage fusion strategy ensures accurate and harmonious camera cloning.

\subsubsection{The effect of Adaptive CFG.}
The row labeled ``w/o AdaCFG'' in the table presents the results of replacing adaptive CFG with a full-stage injection of the camera motion signal. As shown, this leads to a significant drop in camera accuracy. Furthermore, the visual results in Figure \ref{fig:ablation_vis} demonstrate that without AdaCFG, the camera rotates noticeably more slowly. This occurs because the model is forced to process multiple signals simultaneously, resulting in an insufficient response to the camera motion guidance.
\begin{figure}
\begin{center}
\includegraphics[width=1.0\linewidth]{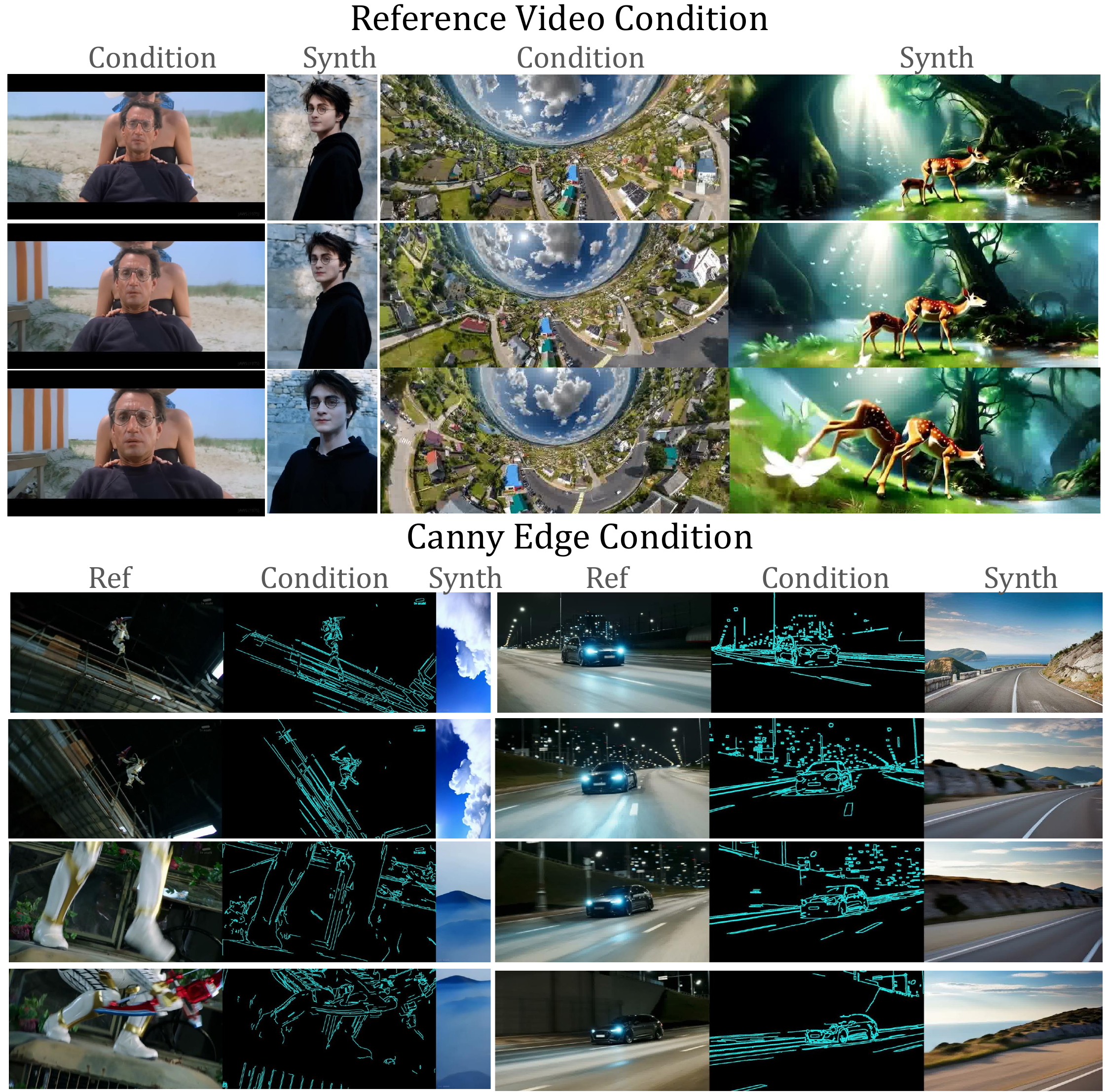}
\end{center}
\caption{\textbf{Emergent zero-shot camera control in OmniDirector}. During inference, substituting the camera grid with raw RGB videos or Canny edge sequences effectively drives camera motion, demonstrating robust generalization without any retraining.}
\label{fig:emergent}
\end{figure}

\subsection{Emergent Camera Understanding}

Fundamentally, the camera grid functions as a visual proxy for spatial movement, sharing highly similar spatiotemporal semantics with RGB videos. Empirically, we discover that this representation effectively unlocks an emergent capability within the video generation model for comprehending camera dynamics. As illustrated in Figure \ref{fig:emergent}, with model parameters frozen during inference, the model robustly deduces and executes camera motions when conditioned on diverse signals, including the raw reference video or a Canny edge video. This strong cross-domain generalization not only mitigates potential errors stemming from inaccurate pose estimation methods \cite{lin2025dpa3, keetha2026mapanything}, but also enables models to clone complex cinematographic techniques—such as the Hitchcock zoom and distortion depicted in the figure without requiring explicitly curated training data for such specific trajectories.

\section{Conclusion}

In this paper, we propose OmniDirector, which achieves general multi-shot camera cloning. By rendering the camera parameters within a 3D empty room, we represent the camera motion as grid videos. Based on this, we construct million-scale camera grid-video training pairs to empower OmniDirector with camera control. Furthermore, we design a Hierarchical Prompt Expansion Agent during inference to harmoniously integrate camera motion with other multimodal control signals. Overall, OmniDirector provides an efficient and accessible paradigm for multi-shot camera cloning.

\noindent\textbf{Limitations and Future Work.}  
OmniDirector currently employs direct token concatenation to integrate the multimodal control signals, which struggles to maintain long-term memory and temporal consistency when scaling to significantly longer video sequences. 
In future work, we plan to explore advanced temporal memory mechanisms (such as long-context cross-attention modules or memory banks) to enhance the model's ability.

\section*{Acknowledgments}
\bigskip
\noindent We sincerely thank Mingyang Shan, Fanqi Meng, Wanqi Shi and Jiaxin Hu for contributing to the evaluation part.
\bibliography{aaai2027}

\end{document}